\documentclass[preprint,10pt]{elsarticle}

\usepackage{amssymb}
\usepackage{graphicx}
\usepackage{multicol}
\usepackage{balance}
\usepackage{stfloats}
\usepackage[cmex10]{amsmath}
\usepackage{amsmath}
\usepackage{color}
\usepackage{CJK}
\usepackage{algorithm}
\usepackage{algorithmicx}
\usepackage{algpseudocode}

\begin{document}

\begin{frontmatter}



\title{Low rank prior and $l_0$ norm to remove impulse noise in images}


\author{Haijuan Hu}
\address{Northeastern University at Qinhuangdao,
              School of Mathematics and Statistics,  Hebei, 066004, China.
              ({huhaijuan61@126.com}).}


\begin{abstract}
Patch-based low rank is an important prior assumption for image processing.
 Moreover, according to our calculation, the optimization of $l_0$ norm corresponds to the maximum likelihood estimation under random-valued impulse noise. In this article, we thus combine exact rank and $l_0$ norm  for removing the noise. It is solved formally using the alternating
direction method of multipliers (ADMM), with our previous patch-based weighted filter (PWMF) producing initial images. Since this model is not convex, we consider it as a Plug-and-Play ADMM, and do not discuss theoretical
convergence properties. Experiments show that this method has very good performance, especially for weak or medium contrast images.
\end{abstract}


\end{frontmatter}




\section{Introduction}
Image restoration is a fundamental problem in image processing. We focus on here random-valued impulse noise, often
 added during image acquisition and transmission. The task is to recover original image from the noisy one by removing noise.

To recover images, one should assume images have some essential properties. Traditional prior  assumptions are smoothness\cite{hunt1973application}, piecewise constant\cite{rudin_nonlinear_1992}, etc. Among many priors developed later, two important priors are  patch similarity\cite{buades_review_2006}, and low rank property \cite{gu2014weighted, hu2018note}.
Patch similarity assumes that there exist many similar patches in images. Low rank property means the matrix composed of similar patches has low rank.    Based on patch similarity and a statistics ROAD\cite{garnett2005universal} for detecting impulse noise, Hu et al\cite{hu2016removing} propose patch-based weighted filter (PWMF) for removing impulse noise and mixed noise.
 Since the optimization of the rank is generally a NP-hard problem,
 many researches use the nuclear norm or the weighted nuclear norm to approximate the rank, such as WNNM\cite{gu2014weighted} for removing Gaussian noise. But in some cases, the rank optimization has a close-form solution:  Hu et al\cite{hu2018note} propose an exact low rank model for removing Gaussian noise. In \cite{jin2018sparse}, the authors use the sparse and low rank decomposition of a Hankel structured matrix for impulse noise removal, where $l_1$ norm is used for the sparseness, and nuclear norm for approximating rank. In \cite{adam2021combined}, the authors propose a variational model that uses the nonconvex
$l_p$-norm, $0<p<1$ for both the data fidelity and a regularization term, combined with a second-order total variation regularization
term and an overlapping group sparse regularizer. Recently deep learning is also applied for removing impulse noise, and has reached very good performance\cite{jin2019learning, wang2020variational}.

However, to remove impulse noise, the exact low rank properties are not exploited in the literature as far as we know. In this paper,  we will extend the exact low rank model\cite{hu2018note} to remove impulse noise. Moreover, according to our computation, $l_0$ norm fidelity corresponds to the maximum likelihood estimate under impulse noise, noting that $l_2$ norm corresponds to Gaussian noise. Thus we optimize a combination of exact rank and $l_0$ norm
fidelity, which is solved formally by the alternating
direction method of multipliers (ADMM) \cite{goldstein2009split}. We consider the method as a Plug-and-Play ADMM\cite{venkatakrishnan2013plug}, and do not discuss theoretical
convergence properties. Since the model is not convex, the results depend on the initial image. We choose the PWMF results as initial images, which can also simplify the model. Experiments show that this algorithm produces very good images, especially for low or medium contrast images.

\section{Patch-based low rank method } 
\label{sec_plr}
In \cite{hu2018note}, Hu et al propose a patch-based low rank method (PLR) for removing Gaussian noisse. Firstly, the noisy image is divided into overlapping patches of size $d\times d$. For each patch, its $m$ most similar patches located in its $M\times M$ neighbor region are grouped together to form a similarity matrix, each column being a vectorization of a similar patch. The similarity is determined by Euclidean norm. For each similarity matrix $\boldsymbol S$, it is denoised by the exact low rank model
\begin{equation}
\min_X \|\boldsymbol S-X\|_F^2+mt^2 \mbox{Rank} (X),
\label{eq_lowrank}
\end{equation}
where the minimum is taken over all the matrices $X$ having the same size as $\boldsymbol S$, $\|\cdot\|_F$ is the Frobenius norm, and $t^2$ represents a threshold parameter($t^2\approx 2\sigma^2$).
The solution can be obtained by
\begin{equation}
\hat{\boldsymbol S}:=\boldsymbol P H_{t\sqrt{m}}(\boldsymbol \Sigma) \boldsymbol Q^T,
\label{eq_lr_ans2}
\end{equation}
 where $\boldsymbol P$, $\boldsymbol Q$, and $\boldsymbol \Sigma$ are derived from the SVD of $\boldsymbol S$, that is, $\boldsymbol S=\boldsymbol P\boldsymbol \Sigma  \boldsymbol Q^T$, and $H_{t\sqrt{m}}$ is the hard thresholding operator:
\begin{equation} H_{t\sqrt{m}}(\boldsymbol \Sigma)_{kk}=\left\{\begin{array}{rl}
\boldsymbol \Sigma_{kk} & \mbox{if} \;\; \boldsymbol \Sigma_{kk}> t\sqrt{m}, \\
0 & \mbox{otherwise,}\end{array}  \quad k=1,2,\cdots, d^2. \right.
\label{eq_hardsvd}
\end{equation}
Finally, each denoised similarity  matrix  is returned to the original location; the overlapping denoised values for each pixel are aggregated. The parameters are chosen as $d=7, M=43, m=245, t=1.5\sigma$.


\section{Remove impulse noise}
The noise model is as follows,
\begin{equation} \boldsymbol v_0=\left\{\begin{array}{ll}
\boldsymbol\eta & \mbox{with probalility }p \\
\boldsymbol u_0 & \mbox{with probalility} 1-p \end{array} \right.
\label{noise}
\end{equation}
where $\boldsymbol u_0 $ is the original image, $\boldsymbol v_0 $ is the observed noisy one, $\boldsymbol \eta$ is independent random noise uniformly distributed on [min$\{\boldsymbol u_0\}$, max$\{\boldsymbol u_0\}]$, generally taken as [0,255] for 8-bit gray images, and $p (0<p<1)$ is the proportion of noise, representing the noise level.

As it is known that the fidelity term for Gaussian noise is $l_2$ norm, to obtain the fidelity term for impulse noise,  we consider the maximum likelihood estimation (MLE) for $\boldsymbol u_0$. For simplicity, we utilize the discrete model. That is, for two pixels $u$ and $v$ of the same location of $\boldsymbol u_0$ and $\boldsymbol v$ respectively,
\begin{equation} P\{ v=k\}=\left\{\begin{array}{ll}
\frac p {256} & k\in\{0,1,2\cdots 255\}/{u}, \\
1-p+\frac p {256} & k\in \{u\}. \end{array}  \right.
\label{noise}
\end{equation}
Given $n$ independent realizations $\{v_1,v_2,\cdots, v_n\}$ of $v$, the likelihood function is then
\begin{equation}
L(v_1,v_2,\cdots, v_n)=(1-p+\frac p {256})^N(\frac p {256})^{n-N},
\end{equation}
where $N=n-\sum_i^n\|v_i-u\|_0$ is the number $v_i$ taking the value $u$. Recall that
\begin{equation}\|v_i-u\|_0=\left\{\begin{array}{cl}
0 & v_i=u, \\
1 & v_i \not= u. \end{array} i=1,2,\cdots,n \right.
\label{noise}
\end{equation}
Since $1-p+\frac p {256}>\frac p {256}$, $L(v_1,v_2,\cdots, v_n)$ is an increasing function of $N$. That is,
\begin{equation}
\arg\max_u L(v_1,v_2,\cdots, v_n)=\arg\max_u N=\arg\min_u \sum_i^n\|v_i-u\|_0,
\end{equation}
which means the fidelity term for impulse noise is $l_0$ norm.

According to above analysis, the following model is considered to remove impulse noise,
\begin{equation}
 \underset{\boldsymbol u}{\mbox{min}}\; \mu\sum_l Rank(R_l\boldsymbol u) +\|\boldsymbol u-\boldsymbol v_0\|_0, \label{eq_model_mixed}
\end{equation}
where $R_l\boldsymbol u$ denotes similarity matrices.
 Since the model is not convex,  it is solved formally by ADMM \cite{goldstein2009split,ma2017low}.
\begin{eqnarray}
 &   \boldsymbol u^{k+1}&=\mbox{arg}\underset{\boldsymbol u}{\mbox{min}}\;  \|(\boldsymbol u-\boldsymbol v_0)\|_0+\frac{\alpha}2\|\boldsymbol u-\boldsymbol v^k-\boldsymbol b^k\|_F^2, \label{usplit_mixed}\\
 & \boldsymbol v^{k+1}&=\mbox{arg}\underset{\boldsymbol v}{\mbox{min}}\; \mu\sum_l Rank(R_l \boldsymbol v) +\frac{\alpha}2\|\boldsymbol u^{k+1}-\boldsymbol v-\boldsymbol b^k\|_F^2,\label{vsplit_mixed}\\
 & \boldsymbol b^{k+1}&=\boldsymbol b^k+\boldsymbol v^{k+1}-\boldsymbol u^{k+1}.
  \end{eqnarray}
 We use  PWMF to produce initial image $\boldsymbol v^0$ for the iteration, since it runs fast and has good results; the initial image $\boldsymbol b^0$ is the zero matrix.
 By an element-wise calculation, the minimizer of (\ref{usplit_mixed}) has a close form:
 \begin{equation}
   \boldsymbol u^{k+1}=
   \left\{\begin{array}{l}
   \boldsymbol v_0,\;\;\;\; \;\;\;\;\mbox{if} \; |\boldsymbol v^k+b^k-\boldsymbol v_0|<\sqrt{2/\alpha},\\
   \boldsymbol v^k+\boldsymbol b^k,\; \mbox{else}.
   \end{array}
   \right.
    \end{equation}
The $v$-subproblem (\ref{vsplit_mixed}) is solved as PLR in Section \ref{sec_plr}, with $\boldsymbol u^{k+1}-\boldsymbol b^k$ regarded as Gaussian noisy images and $mt^2=2\mu/\alpha$.
Since   $\boldsymbol v^0$ contains little impulse noise,  $\boldsymbol u^{k+1}-\boldsymbol b^k$ are good estimations of original images, which are thus used directly for the selection of similar patches for $v$-sub problem as PLR.

 \section{Experiments}
\begin{figure}
\begin{center}
\renewcommand{\arraystretch}{0.5} \addtolength{\tabcolsep}{-6pt} \vskip3mm {%
\fontsize{8pt}{\baselineskip}\selectfont

\begin{tabular}{ccccc}
\includegraphics[width=0.15\linewidth]{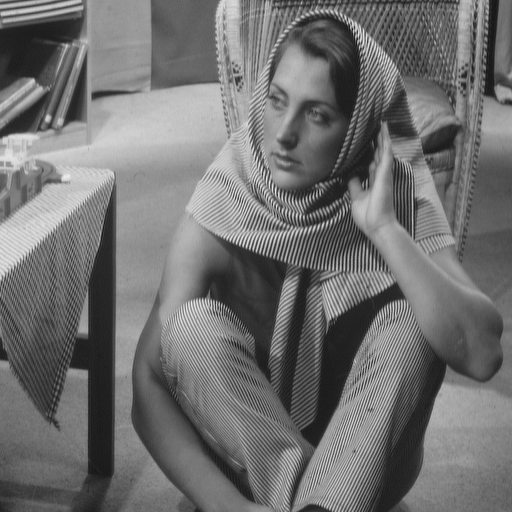}
&\includegraphics[width=0.15\linewidth]{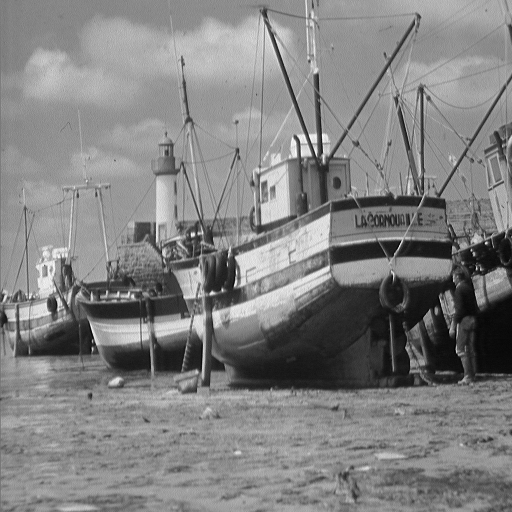}
&\includegraphics[width=0.15\linewidth]{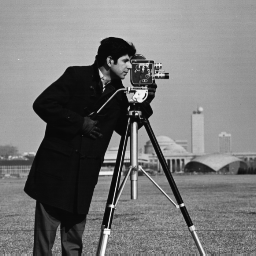}
&\includegraphics[width=0.15\linewidth]{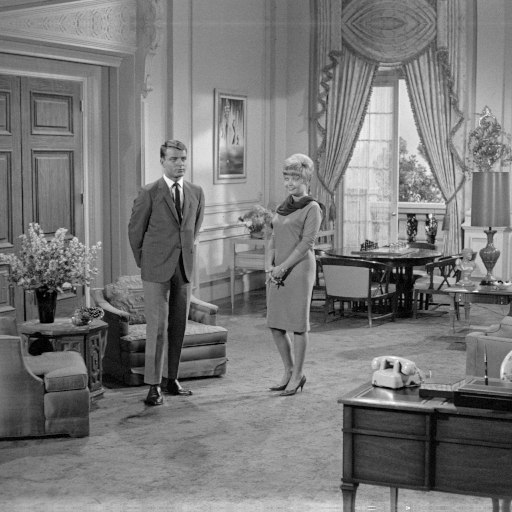}
&\includegraphics[width=0.15\linewidth]{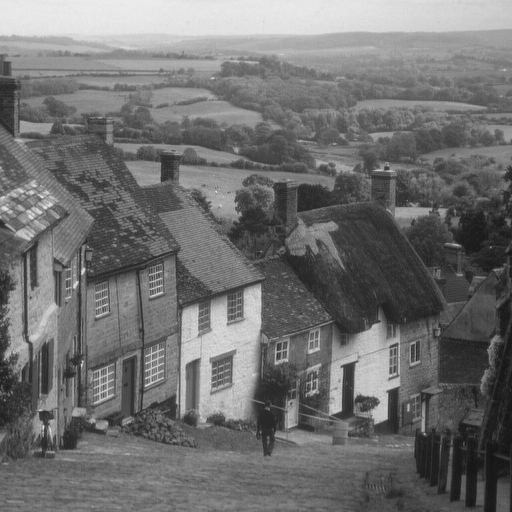} \\
Barbara&Boats&Cameraman& Couple &Hill \\
\includegraphics[width=0.15\linewidth]{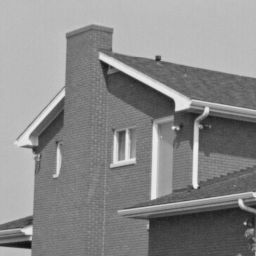}
&\includegraphics[width=0.15\linewidth]{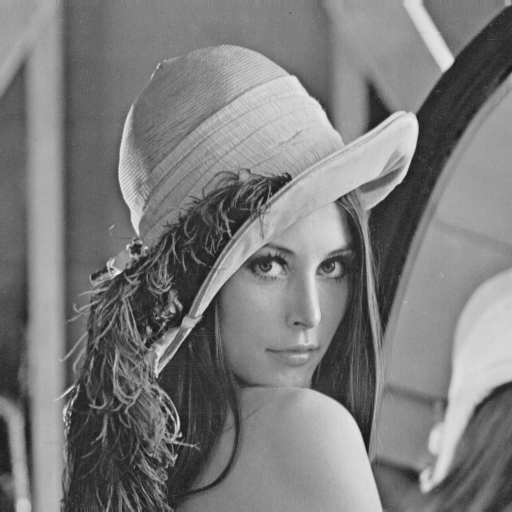}
&\includegraphics[width=0.15\linewidth]{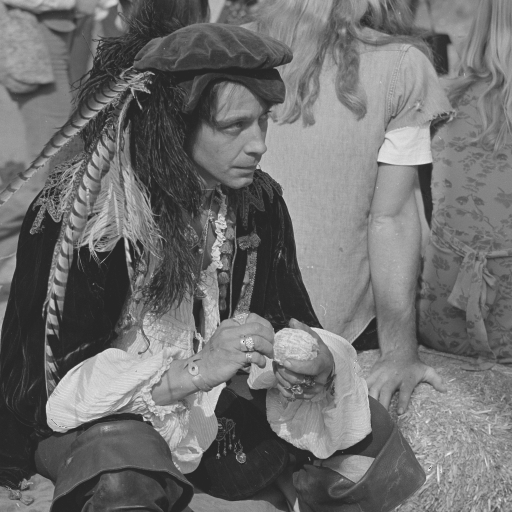}
&\includegraphics[width=0.15\linewidth]{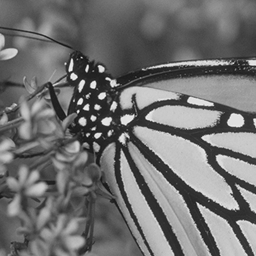}
&\includegraphics[width=0.15\linewidth]{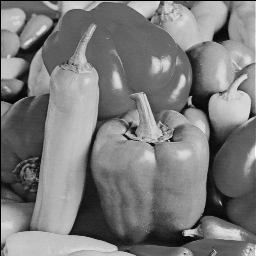}\\
House&Lena & Man&Monarch &Peppers
\end{tabular}
} 
\par
\rule{0pt}{-0.2pt}
\par
\vskip -1mm
\caption{ Original images tested }
\label{orifig}
\end{center}
\end{figure}

To show the performance of the proposed method, we test some classic images, the original clean images of which are displayed in Figure \ref{orifig}. We use peak signal-to-noise ratio (PSNR) values to measure the quality of denoised images $\bar{\boldsymbol v}$ quantitatively, which are defined by
PSNR$(\bar {\boldsymbol v})=20\log_{10}(255r/\|\bar {\boldsymbol v}-\boldsymbol u_0\|_F)$dB with $r$ being the squared root of the number of elements of the image $\boldsymbol u_0$.  For our method, we only need to choose the values of $\mu$ and $\alpha$, and determine the stop criteria for the iteration. We use PLR with $t=7.5$ in (\ref{eq_lowrank}) for the solution of (\ref{vsplit_mixed}) assuming $\boldsymbol u^{k+1}-\boldsymbol b^k$ has  Gaussian noise level $\sigma=5$. We choose $\alpha$ empirically by obtaining the highest average PSNR values of tested images, which is taken as $\alpha=1/72$ for all the impulse noise levels $p=0.2,0.3,0.4,0.5$. Comparing (\ref{vsplit_mixed}) and (\ref{eq_lowrank}), we thus use $\mu=mt^2\alpha/2\approx 95.7$. In addition, the fixed iteration times 50 is used.

The proposed method is compared with some recently published methods, robust ALOHA\cite{jin2018sparse}, HNHOTV-OGS\cite{adam2021combined}, and our previous method PWMF\cite{hu2016removing}. We consider four noise level with $p=0.2, 0.3, 0.4$, and 0.5. For PWMF, we use the same parameters as in the paper\cite{hu2016removing}. For robust ALOHA, since the paper\cite{jin2018sparse} only considers $p=0.25$ and $p=0.4$, we use the same parameters as $p=0.25$ when $p=0.2$ and 0.3, and  the same parameters as $p=0.4$ when $p=0.4$ and 0.5. For
 HNHOTV-OGS\cite{adam2021combined}, according to the paper, we adjust the parameters $\lambda$ in [43,79], and $\omega$ in [4.8, 7.5],which are finally  chosen as $\lambda=77, \omega=6.2$ for $p=0.2$; $\lambda=77, \omega=6.4$ for $p=0.3$; $\lambda=75, \omega=6.8$ for $p=0.4$, and $\lambda=75, \omega=7.2$ for $p=0.5$ by obtaining highest average PSNR values of tested images.

\begin{figure}
\begin{center}
\renewcommand{\arraystretch}{0.5} \addtolength{\tabcolsep}{-6pt} \vskip3mm {%
\fontsize{8pt}{\baselineskip}\selectfont
\begin{tabular}{ccccccc}
\includegraphics[width=0.3\linewidth]{Peppers.png}
&\includegraphics[width=0.3\linewidth]{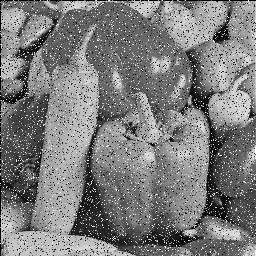}
&\includegraphics[width=0.3\linewidth]{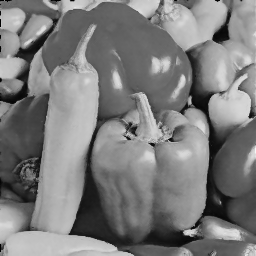}\\
Original&Noisy&PWMF\\
\includegraphics[width=0.3\linewidth]{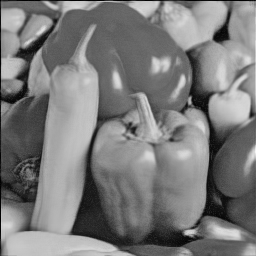}
&\includegraphics[width=0.3\linewidth]{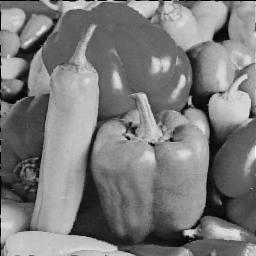}
&\includegraphics[width=0.3\linewidth]{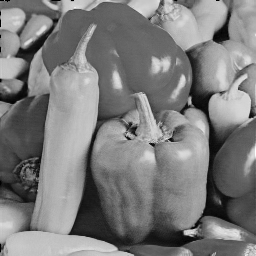}\\
Robust ALOHA & HNHOTV-OGS &Proposed\\
\includegraphics[width=0.3\linewidth,trim=80 110 100 70,clip]{Peppers.png}
&\includegraphics[width=0.3\linewidth,trim=80 110 100 70,clip]{noisy_Peppers_p2.png}
&\includegraphics[width=0.3\linewidth,trim=80 110 100 70,clip]{PWMF_Peppers_p=02.png}
\\
\includegraphics[width=0.3\linewidth,trim=80 110 100 70,clip]{ALOHA_peppers_p=02.png}
&\includegraphics[width=0.3\linewidth,trim=70 110 110 70,clip]{ADAME_Peppers_p=02.png}
&\includegraphics[width=0.3\linewidth,trim=80 110 100 70,clip]{MY_Peppers_p=02.png}\\
\end{tabular}
} 
\par
\rule{0pt}{-0.2pt}
\par
\vskip -1mm
\caption{Denoised images for $p=0.2$. The third and bottom rows show the enlarged versions of the same part of the Peppers images from the first and second rows respectively. }
\label{p2figPepper}
\end{center}
\end{figure}

\begin{figure}
\begin{center}
\renewcommand{\arraystretch}{0.5} \addtolength{\tabcolsep}{-6pt} \vskip3mm {%
\fontsize{8pt}{\baselineskip}\selectfont

\begin{tabular}{ccccccc}
\includegraphics[width=0.3\linewidth]{House.png}
&\includegraphics[width=0.3\linewidth]{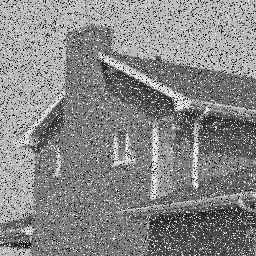}
&\includegraphics[width=0.3\linewidth]{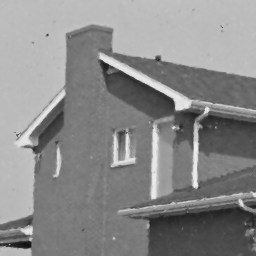}\\
Original&Noisy&PWMF\\
\includegraphics[width=0.3\linewidth]{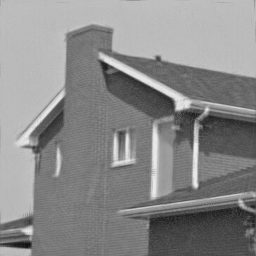}
&\includegraphics[width=0.3\linewidth]{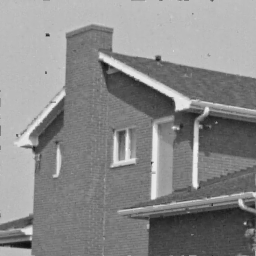}
&\includegraphics[width=0.3\linewidth]{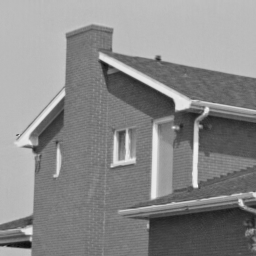}\\
Robust ALOHA & HNHOTV-OGS &Proposed\\
\includegraphics[width=0.3\linewidth,trim=147 73 59 133,clip]{House.png}
&\includegraphics[width=0.3\linewidth,trim=147 73 59 133,clip]{noisy_House_p3.png}
&\includegraphics[width=0.3\linewidth,trim=147 73 59 133,clip]{PWMF_House_p=03.png}\\
\includegraphics[width=0.3\linewidth,trim=147 73 59 133,clip]{ALOHA_house_p=03.png}
&\includegraphics[width=0.3\linewidth,trim=147 73 59 133,clip]{ADAME_House_p=03.png}
&\includegraphics[width=0.3\linewidth,trim=147 73 59 133,clip]{MY_House_p=03.png}\\

\end{tabular}
} 
\par
\rule{0pt}{-0.2pt}
\par
\vskip -1mm
\caption{Denoised images for $p=0.3$ The third and bottom rows show the enlarged versions of the same part of the House images from the first and second rows respectively.}
\label{p3fig}
\end{center}
\end{figure}

\begin{figure}
\begin{center}
\renewcommand{\arraystretch}{0.5} \addtolength{\tabcolsep}{-6pt} \vskip3mm {%
\fontsize{8pt}{\baselineskip}\selectfont

\begin{tabular}{ccccccc}
\includegraphics[width=0.3\linewidth]{Lena.png}
&\includegraphics[width=0.3\linewidth]{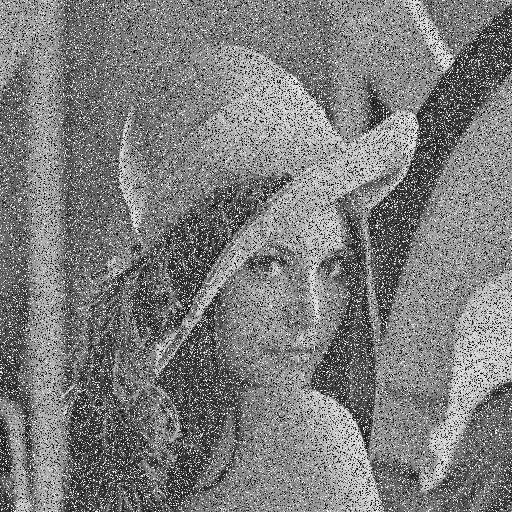}
&\includegraphics[width=0.3\linewidth]{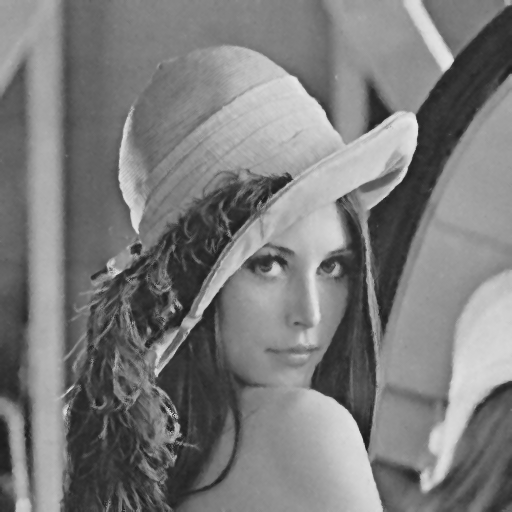}\\
Original&Noisy&PWMF\\
\includegraphics[width=0.3\linewidth]{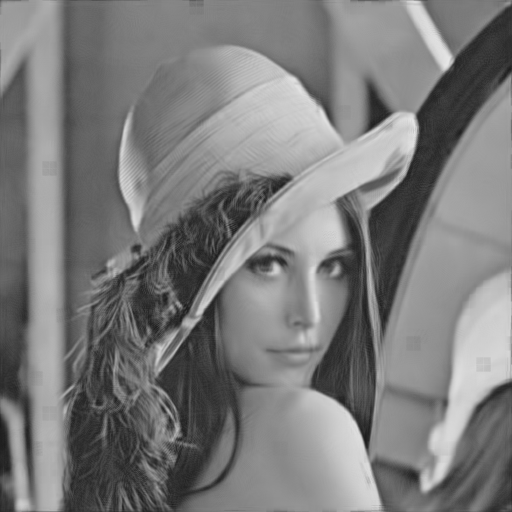}
&\includegraphics[width=0.3\linewidth]{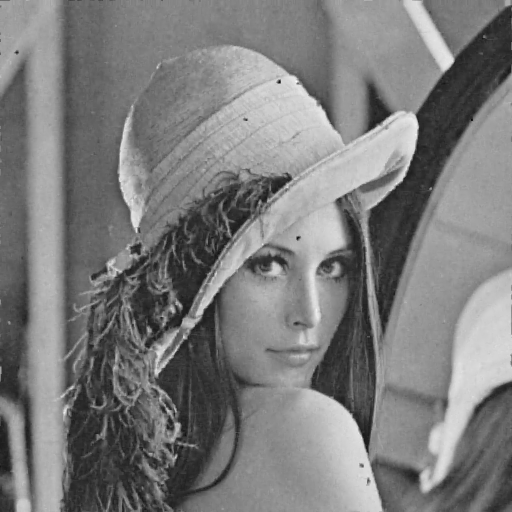}
&\includegraphics[width=0.3\linewidth]{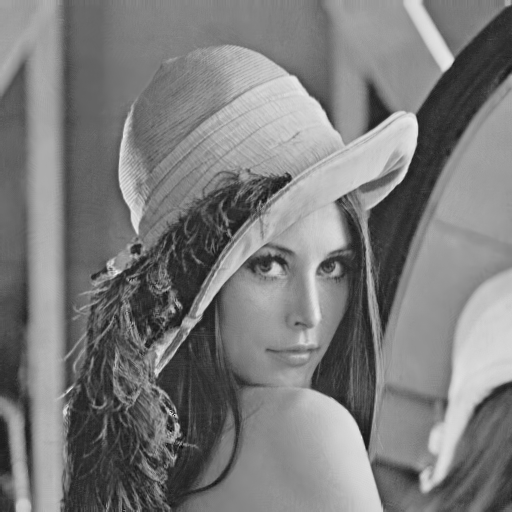}\\
Robust ALOHA & HNHOTV-OGS &Proposed\\
\includegraphics[width=0.3\linewidth,trim=270 322 162 110,clip]{Lena.png}
&\includegraphics[width=0.3\linewidth,trim=270 322 162 110,clip]{noisy_Lena_p4.png}
&\includegraphics[width=0.3\linewidth,trim=270 322 162 110,clip]{PWMF_Lena_p=04.png}\\
\includegraphics[width=0.3\linewidth,trim=270 322 162 110,clip]{ALOHA_lena_p=04.png}
&\includegraphics[width=0.3\linewidth,trim=270 322 162 110,clip]{ADAME_Lena_p=04.png}
&\includegraphics[width=0.3\linewidth,trim=270 322 162 110,clip]{MY_Lena_p=04.png}\\
\end{tabular}
} 
\par
\rule{0pt}{-0.2pt}
\par
\vskip -1mm
\caption{Denoised images for $p=0.4$. The third and bottom rows show the enlarged versions of the same part of the Lena images from the first and second rows respectively.}
\label{p4fig}
\end{center}
\end{figure}

\begin{figure}
\begin{center}
\renewcommand{\arraystretch}{0.5} \addtolength{\tabcolsep}{-6pt} \vskip3mm {%
\fontsize{8pt}{\baselineskip}\selectfont
\begin{tabular}{ccccccc}
\includegraphics[width=0.3\linewidth]{Boats.png}
&\includegraphics[width=0.3\linewidth]{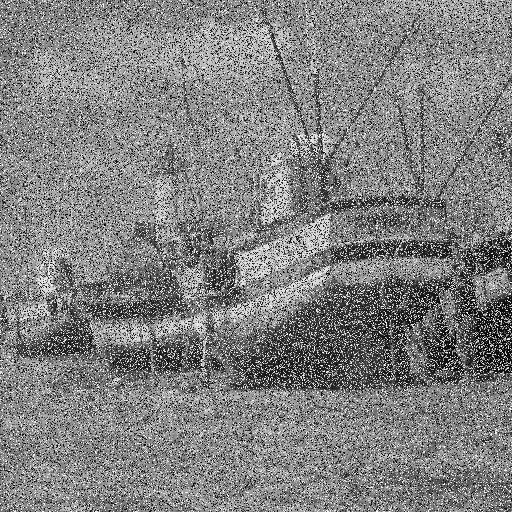}
&\includegraphics[width=0.3\linewidth]{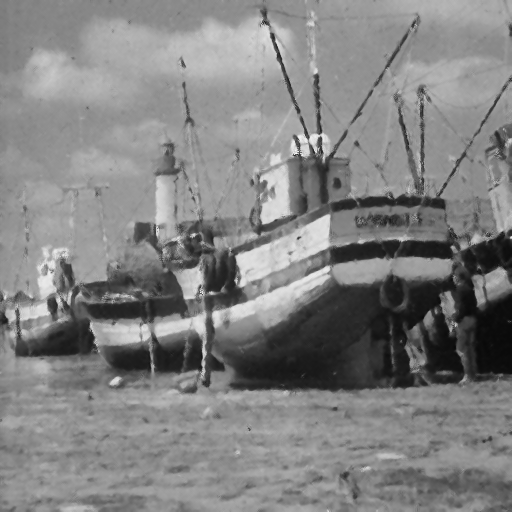}\\
Original&Noisy&PWMF\\
\includegraphics[width=0.3\linewidth]{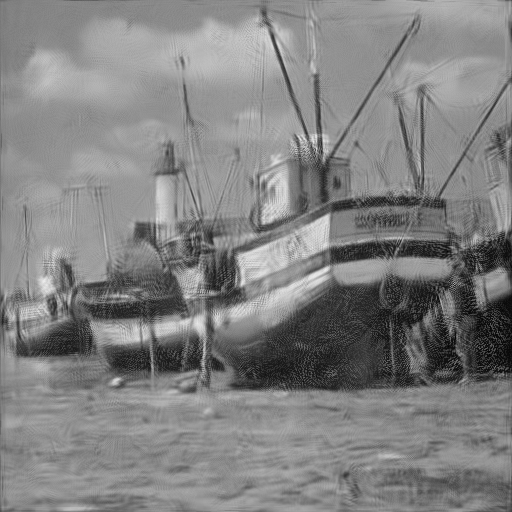}
&\includegraphics[width=0.3\linewidth]{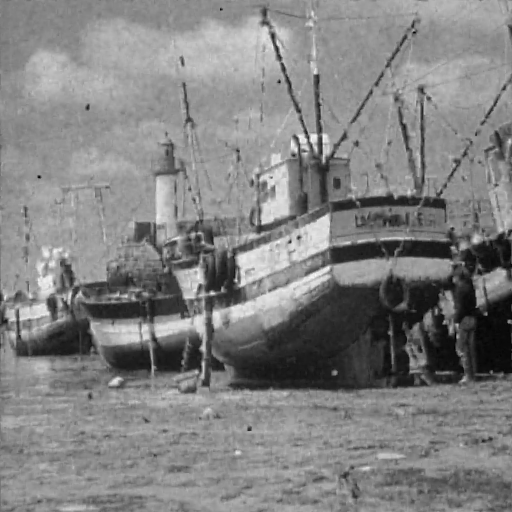}
&\includegraphics[width=0.3\linewidth]{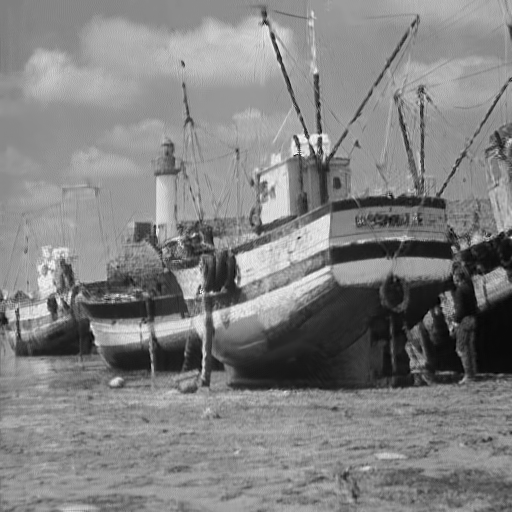}\\
Robust ALOHA & HNHOTV-OGS &Proposed\\
\includegraphics[width=0.3\linewidth,trim=150 262 262 150,clip]{Boats.png}
&\includegraphics[width=0.3\linewidth,trim=150 262 262 150,clip]{noisy_Boats_p5.png}
&\includegraphics[width=0.3\linewidth,trim=150 262 262 150,clip]{PWMF_Boats_p=05.png}\\
\includegraphics[width=0.3\linewidth,trim=150 262 262 150,clip]{ALOHA_boat_p=05.png}
&\includegraphics[width=0.3\linewidth,trim=150 262 262 150,clip]{ADAME_Boats_p=05.png}
&\includegraphics[width=0.3\linewidth,trim=150 262 262 150,clip]{MY_Boats_p=05.png}\\

\end{tabular}
} 
\par
\rule{0pt}{-0.2pt}
\par
\vskip -1mm
\caption{Denoised images for $p=0.5$. The third and bottom rows show the enlarged versions of the same part of the Boats images from the first and second rows respectively.}
\label{p5fig}
\end{center}
\end{figure}

The PSNR values of different methods are shown in Table \ref{tab_impnoise1}. It is shown that our method is better than other methods for most of the images,
 especially for images with weak contrast, such as Lena and House. Robust ALOHA\cite{jin2018sparse} is especially good for the high contrast image Barbara and is better for the image Cameraman when $p=0.2$ than other methods, but it is not as good as HNHOTV-OGS\cite{adam2021combined} or the proposed method in other cases. HNHOTV-OGS is better than the proposed method in only three cases, and the differences are small. The proposed method improves PWMF in all cases.
In Figures \ref{p2figPepper},\ref{p3fig},\ref{p4fig}, and \ref{p5fig}, some images are displayed to show the differences of the methods. From these figures, it can be seen that our method improves PWMF: it retains clear image details in textural parts and edges, and also recovers well homogeneous regions. From Figures \ref{p2figPepper} and \ref{p4fig}, it can be observed that robust ALOHA loses more details than our methods, while  some noise or artifacts appear in Figures \ref{p3fig} and \ref{p5fig}.
The image details obtained by HNHOTV-OGS are not as clear as the proposed method either, which can be seen in Figures \ref{p3fig}, \ref{p4fig}, and \ref{p5fig}.

 Note that we use the same parameters for all the tested images. Since different images have different structures, the denoising performance can be improved with adaptive parameters, especially for high contrast images, which will be our future work.

 \begin{table}
\begin{center} 
\caption{PSNR values for removing impulse noise with $p=0.2, 0.3, 0.4$ and 0.5. For each $p$, from top to bottom, the methods are PWMF\cite{hu2016removing}, robust ALOHA\cite{jin2018sparse}, HNHOTV-OGS\cite{adam2021combined}, and the proposed method. From left to right, the images are Barbara, Boats, Cameraman, Couple, Hill, House, Lena, Man, Monarch, Peppers.}
\begin{tabular}{ccccccccccccccccc}
\hline
\noalign{\smallskip}
 Bar &Boa&Cam&Cou&Hill&Hou & Lena&Man&Mon&Pep \\
\noalign{\smallskip}
\hline
\noalign{\smallskip}%
$p=0.2 $\\
\hline
 28.43    & 31.86   & 27.26   & 31.94    &   33.59    &  35.06   &  35.77   &  33.07   &  28.63   &  32.08 \\
\textbf{34.79}&  31.05&   \textbf{28.09}&  31.06&  32.65& 34.57&  35.36&  31.15&   28.29 &30.10\\
25.83&   32.49&  27.13&   33.36&   33.84&   37.24&   36.48&  32.86&   \textbf{30.29}&   31.87\\
 29.46  & \textbf{33.68}  &  27.68   &   \textbf{33.51}  & \textbf{36.26}   &  \textbf{40.37}   &  \textbf{38.83}    & \textbf{34.11}   & 29.97    &  \textbf{33.17}  \\
\noalign{\smallskip}\hline
$p=0.3 $\\
\hline
   26.33& 29.76&   25.67&  29.69&   31.16&   32.70&   33.61&   30.88&   26.93&   29.84\\
      \textbf{31.97}&   29.59&   24.42&   28.49&   29.21&   33.05&   32.74&   29.63&   26.23 &28.19\\
      24.62&   30.31&   25.45&   30.74&   31.28&   33.37&   33.69&   30.87&   27.56&   29.87\\
      27.1&   \textbf{31.33}&  \textbf{25.95}&  \textbf{31.20}&   \textbf{33.73}&  \textbf{37.08}&   \textbf{36.38}&  \textbf{ 32.03}&   \textbf{28.10}&   \textbf{30.84}\\
  \noalign{\smallskip}\hline
  $p=0.4 $\\
  \hline
    25.09&   27.77&   23.74&  27.65&   29.83&   30.93&  31.81&   29.31&  25.06&  28.03\\
   \textbf{29.30}&   26.84&   23.53&   26.91&   28.59&   30.27&   30.11&   27.24&   23.25&  26.24\\
   23.60& 28.28&  \textbf{23.88}&  \textbf{28.69}& 29.19& 30.57& 31.39&29.12& 25.08& 28.16\\

  25.64&   \textbf{28.59}&   23.76&   28.62&  \textbf{32.07}&   \textbf{33.75}&    \textbf{33.46}&    \textbf{29.91}&    \textbf{25.62}&   \textbf{28.66}\\
  \noalign{\smallskip}\hline
 $ p=0.5 $\\
  \hline
  24.06&  26.57&   23.01&  26.38&   28.52&   29.40&   30.38&   27.99&   23.67&    26.36\\
  \textbf{25.61}&  24.21&   20.73&   24.63&  25.36&   26.88&   27.70&   25.65&   21.3&  23.48\\
  23.00& 26.34& 22.66& 26.48& 27.26& 28.77& 29.50& 27.32&  22.99& 26.34\\
24.53&   \textbf{27.07}&  \textbf{ 23.04}&  \textbf{ 27.08}&  \textbf{ 30.12}&  \textbf{31.57}&   \textbf{31.75}&   \textbf{28.42}&   \textbf{24.15}&   \textbf{26.94}\\
    \noalign{\smallskip}\hline
 \end{tabular}

\label{tab_impnoise1}
\end{center}
\end{table}

\section*{References}





\end{document}